\documentclass[letterpaper, 10 pt, journal]{IEEEtran}
\IEEEoverridecommandlockouts                          
\usepackage{graphics} 
\usepackage{epsfig}
\usepackage{times}
\usepackage{amsmath}
\usepackage{amssymb}
\usepackage{threeparttable}
\usepackage{booktabs}
\usepackage{soul}
\usepackage{subfigure}
\usepackage{xcolor}
\usepackage[hidelinks]{hyperref}
\usepackage{algorithm}
\usepackage{algorithmic,comment}
\usepackage{tcolorbox}

\title{\LARGE \bf
Warm-Starting Optimization-Based Motion Planning for Robotic Manipulators via Point Cloud-Conditioned Flow Matching
}

\author{Sibo Tian$^{1}$, Minghui Zheng$^{1,*}$, and Xiao Liang$^{2,*}$
\thanks{This work was supported by the USA National Science Foundation under Grant No. 2026533/2422826. Portions of this research were conducted with the advanced computing resources provided by Texas A\&M High Performance Research Computing.}
\thanks{$^{1}$ Sibo Tian and Minghui Zheng are with the J. Mike Walker '66 Department of Mechanical Engineering, Texas A\&M University, College Station, TX 77843, USA. {\tt\small Emails: {sibotian, mhzheng}@tamu.edu.}}
\thanks{$^{2}$ Xiao Liang is with the Zachry Department of Civil and Environmental Engineering, Texas A\&M University, College Station, TX 77843, USA. {\tt\small Email: xliang@tamu.edu.}}
\thanks{$^{*}$ Corresponding Authors.}}

\begin{document}

\maketitle
\thispagestyle{empty}
\pagestyle{empty}

\begin{abstract}

Rapid robot motion generation is critical in Human-Robot Collaboration (HRC) systems, as robots need to respond to dynamic environments in real time by continuously observing their surroundings and replanning their motions to ensure both safe interactions and efficient task execution. Current sampling-based motion planners face challenges in scaling to high-dimensional configuration spaces and often require post-processing to interpolate and smooth the generated paths, resulting in time inefficiency in complex environments. Optimization-based planners, on the other hand, can incorporate multiple constraints and generate smooth trajectories directly, making them potentially more time-efficient. However, optimization-based planners are sensitive to initialization and may get stuck in local minima. In this work, we present a novel learning-based method that utilizes a Flow Matching model conditioned on a single-view point cloud to learn near-optimal solutions for optimization initialization. Our method does not require prior knowledge of the environment, such as obstacle locations and geometries, and can generate feasible trajectories directly from single-view depth camera input. Simulation studies on a UR5e robotic manipulator in cluttered workspaces demonstrate that the proposed generative initializer achieves a high success rate on its own, significantly improves the success rate of trajectory optimization compared with traditional and learning-based benchmark initializers, requires fewer optimization iterations, and exhibits strong generalization to unseen environments.

\end{abstract}

\section{Introduction}

\begin{figure}
    \begin{center}
        \includegraphics[width=1.0\columnwidth]{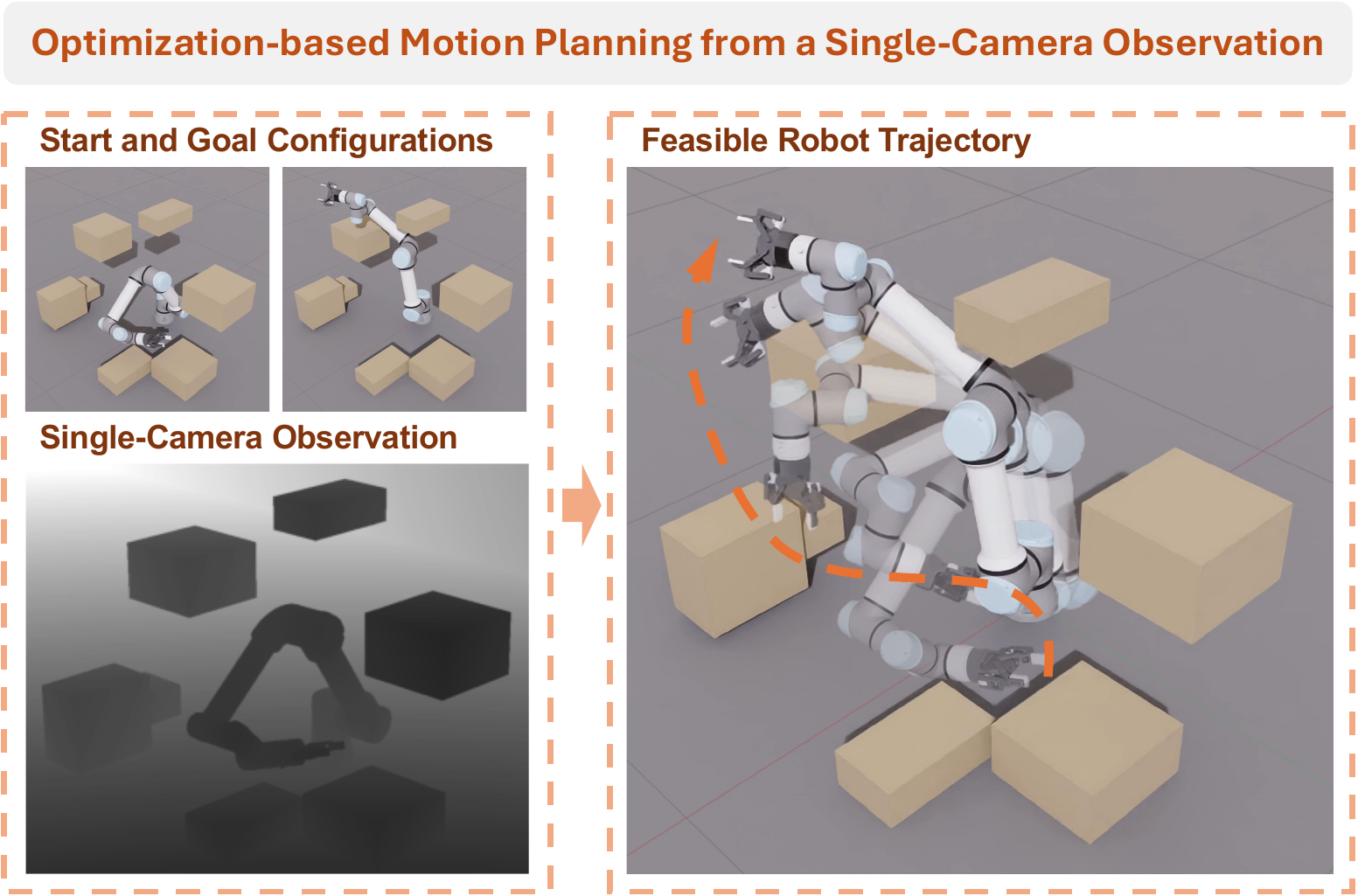}
    \vspace{-0.2in}
    \caption{We present a learning-enhanced trajectory optimization framework that leverages a novel Flow Matching-based initializer to generate multiple diverse seeds for fast and robust optimization. The proposed method requires no prior knowledge of obstacle locations or geometries and can produce smooth, collision-free robot trajectories from a single depth camera observation.}
    \label{overview}
    \end{center}
    \vspace{-0.2in}
\end{figure}

Human-Robot Collaboration (HRC) plays a crucial role in future remanufacturing systems, where humans and robots can work seamlessly together to disassemble and recycle end-of-life products \cite{lee2024review}. Robots contribute strength, precision, and repeatability \cite{liu2025raise}, whereas humans provide perceptual acuity, dexterity, and decision-making capabilities in unstructured tasks. By combining these complementary capabilities, HRC can handle diverse and unpredictable industrial disassembly processes and offer significant advantages over traditional labor-intensive manual remanufacturing. However, such collaboration introduces significant challenges for robotic systems. Shared workspaces are highly dynamic, as humans move within the environment and the spatial configuration evolves when components are removed or repositioned. To ensure safe operation in these settings, robots must continuously perceive their surroundings, anticipate human actions \cite{tian2024transfusion, eltouny2024tgn, tian2024bayesian}, and replan their motions in real time \cite{liu2024integrating}. Achieving this level of responsiveness necessitates advanced motion planning algorithms that can generate motions rapidly while maintaining smoothness, feasibility, and safety constraints.

Robot manipulator motion planning traditionally relies on two main approaches: sampling-based methods and optimization algorithms. Sampling-based planners, such as probabilistic roadmap (PRM) \cite{kavraki2002probabilistic}, rapidly-exploring random tree (RRT) \cite{lavalle1998rapidly}, and their variants \cite{kuffner2000rrt, gammell2015batch}, explore the robot configuration space by incrementally building a roadmap or tree of collision-free configurations. By inferring from the constructed roadmap or tree, the robot can identify a collision-free path connecting the start and goal configurations. Although these methods are probabilistically complete, they usually suffer from the curse of dimensionality, and the resulting paths typically require post-processing or smoothing before execution. Consequently, their performance can be inefficient for high-dimensional planning problems, particularly in cluttered or constrained spaces, where finding feasible paths may incur substantial computational cost. 

On the other hand, optimization-based planners formulate motion planning as a constrained trajectory optimization problem \cite{zucker2013chomp, kalakrishnan2011stomp, schulman2014motion}, where the objective typically incorporates trajectory smoothness, collision avoidance, and task-specific costs. By iteratively refining the trajectory representation, these methods can directly generate high-quality motions that are smooth, efficient, and adaptable to multiple constraints without requiring post-processing. However, since the objective function is often highly non-convex, optimization-based planners are sensitive to initialization and can become trapped in local minima. A common strategy is to initialize the trajectory using linear interpolation between the start and goal configurations. When this initialization fails to produce a feasible solution, a sampling-based planner can be used to provide a better initial guess, improving feasibility at the cost of significantly increased planning time.

In light of these challenges, there is growing interest in leveraging neural networks, which run efficiently at inference time, to augment traditional motion planning methods. Considering that traditional collision checking is time-consuming, several studies have explored neural collision detectors \cite{zhi2022diffco, liu2022deep} to accelerate the planning process and can be easily combined with other works. Biased sampling techniques have been explored in \cite{ichter2018learning, zhang2018learning, wang2020neural}, where machine learning models guide the sampling distribution toward promising regions of the configuration space, thereby improving efficiency and success rates in complex environments. However, these approaches do not directly learn to generate a solution path and are typically integrated with sampling-based planning algorithms. In contrast, imitation learning-based methods \cite{qureshi2019motion, liu2024kg, soleymanzadeh2025simpnet} aim to produce a complete feasible motion path by mimicking expert demonstrations, but they often require multiple rollout steps to generate a single trajectory. What's more, some works utilize neural networks to warm-start numerical optimization \cite{celestini2024transformer, celestini2025generalizable}. By learning a near-optimal seed, the optimization can converge to a feasible solution in fewer iterations. However, these approaches typically assume that the locations and geometries of obstacles are known, which limits their applicability in dynamic or partially observable settings. Moreover, they rely on a deterministic neural initializer that generates only a single seed at a time, thereby limiting the diversity of candidate solutions and reducing robustness in cluttered environments. A recent work \cite{huang2025diffusionseeder} employs a diffusion model conditioned on depth images as the initializer for trajectory optimization, enabling the generation of robot motions directly from depth observations without requiring prior knowledge of obstacles. However, diffusion models typically rely on a multi-step denoising process to refine the trajectory representation iteratively, which can lead to slow inference speeds if distillation is not applied.

To address these limitations, we introduce a novel method that integrates Flow Matching \cite{lipmanflow} with traditional numerical optimization. Unlike diffusion models, Flow Matching inherently learns an optimal straight-line transition from the normal distribution to the data distribution, thereby requiring fewer generation steps. This benefit allows our neural initializer to maintain fast inference speed even when employing a dense trajectory representation. We condition our Flow Matching-based initializer on a single-view point cloud to incorporate perception information, and the obstacle mesh estimated from the point cloud is used for collision checking during the optimization phase. As a result, our approach eliminates the need for prior obstacle knowledge and can directly generate multiple feasible trajectories from raw depth observation, as seen in Fig. \ref{overview}. Moreover, we utilize a state-of-the-art GPU-accelerated trajectory optimizer \cite{sundaralingam2023curobo} in our work for rapid optimization. In summary, our contributions are as follows:

\begin{itemize}
\item We augment traditional trajectory optimization with a novel Flow Matching-based initial seed generator. The proposed stochastic initializer simultaneously produces multiple diverse, near-optimal seeds in as few as two generation steps, accelerating the optimization process and enabling the optimizer to explore solutions across different data modes.
\item Our method conditions robot trajectory generation on a single-view point cloud for perception, enabling planning in partially observable environments without prior knowledge of obstacle locations and geometries.
\item  Simulations with a UR5e manipulator in cluttered environments demonstrate that our approach significantly improves the success rate of trajectory optimization compared to both traditional and other learning-based initialization strategies.
\end{itemize}

\section{Related Work}

Recent advances in deep learning have motivated researchers to explore neural network-based approaches for robot motion planning, providing data-driven alternatives that complement traditional methods. Among these works, different strategies have emerged, including biased sampling to increase the likelihood of exploring relevant regions, imitation learning to mimic expert behaviors, generative models to synthesize feasible motion candidates, and optimization warm-starts to provide favorable initializations for trajectory optimization.

\subsection{Biased Sampling}

To accelerate sampling-based motion planning algorithms via non-uniform sampling without relying on expert-crafted heuristics, various deep learning techniques have been studied. A conditional variational autoencoder (CVAE) is trained in \cite{ichter2018learning} to learn subspaces of valid or desirable robot configurations, conditioned on the details of the planning problem. The latent space can then be used to generate robot configurations in regions where an optimal solution is likely to lie. A policy network is proposed in \cite{zhang2018learning} for a rejection sampling strategy, where the neural network computes a rejection probability for each randomly drawn sample, allowing unfavorable samples to be discarded with high probability and thereby reducing the number of collision checks. Neural RRT* \cite{wang2020neural} trains a convolutional neural network (CNN) to predict the probability distribution of the optimal path, which is then used to guide the sampling process. However, these learning-based biased sampling methods do not directly generate feasible paths and have to be combined with a sampling-based planner.

\subsection{Imitation Learning}

Rather than learning a sampling distribution, imitation learning has been used to generate complete feasible motion plans by directly learning from expert solutions. MPNet \cite{qureshi2019motion} leverages multi-layer perceptrons (MLPs) to predict the next robot joint configuration along the trajectory toward the goal. KG-Planner \cite{liu2024kg} employs a graph neural network (GNN) to capture spatial dependencies among the robot, static obstacles, and other moving agents. Similarly, SIMPNet \cite{soleymanzadeh2025simpnet} encodes the robot’s kinematic structure as a graph and integrates workspace information into the configuration space via a cross-attention mechanism. By applying dropout during inference, MPNet and SIMPNet introduce stochasticity into the planning process, allowing the algorithms to generate alternative paths if a previous attempt results in a collision.

Despite their strong performance, these methods generate trajectories through multi-step rollouts, predicting only the next joint state rather than the entire trajectory, which increases overall planning time.

\subsection{Generative Models}

The biased sampling and imitation learning-based methods discussed above often require post-processing to interpolate and smooth the generated paths into trajectories suitable for the robot to follow, adding extra computational cost. Some works employ generative models for trajectory-level robot motion generation. Motion Planning Diffusion \cite{carvalho2023motion} trains an unconditional diffusion model over kinematically valid trajectories for robot motion generation and, during the inference denoising stage, leverages a customized cost function to guide sampling toward regions that reduce overall cost, gradually producing trajectories that satisfy task-specific constraints such as collision avoidance and trajectory smoothness. Rather than using a single cost, EDMP \cite{saha2024edmp} employs an ensemble of cost functions and runs a batch of cost-guided diffusion processes in parallel to capture scene variations, thereby improving generalization to different environments. In addition to cost-guided diffusion, Diffuser \cite{janner2022planning} represents trajectories as state-action pairs, allowing the state predictor and controller to be derived simultaneously. However, these methods require full knowledge of the environment to compute the cost function, which limits their applicability in partially observable settings. Moreover, since these models introduce conditions only at the inference stage, they typically require multiple denoising steps to gradually refine the trajectory to satisfy task-specific constraints, which can slow down the planning process.

\subsection{Optimization Warm-start}

Although trajectory optimization can refine an initial guess into a collision-free and dynamically feasible solution for a robot to execute, its performance strongly depends on the quality of the initial seed. Poor initializations may cause convergence to local minima or even failure to produce a feasible solution. To mitigate this limitation, recent works \cite{celestini2024transformer, celestini2025generalizable, huang2025diffusionseeder, yoon2023learning} leverage neural networks to generate favorable warm-starts that enhance both convergence speed and solution quality. A transformer is employed in \cite{celestini2024transformer} to auto-regressively generate an open-loop trajectory for initializing non-convex trajectory optimization in model predictive control. Subsequent work \cite{celestini2025generalizable} extends this approach by enabling the transformer to process comprehensive, multi-modal scene representations. A reinforcement learning policy is proposed in \cite{yoon2023learning} to seed the trajectory optimization for a redundant manipulator. However, these works generate seeds in an auto-regressive manner, which slows down the motion generation process. Moreover, these approaches are deterministic, producing only a single near-optimal seed. While this can mitigate the local minima issue to some extent, relying on a single realization still carries the risk of convergence to local minima. DiffusionSeeder \cite{huang2025diffusionseeder} leverages a generative model to produce multiple diverse seeds; however, due to the inherent limitations of diffusion models, their method requires multiple denoising steps to generate clean seeds, which can slow down the motion generation, especially for a cluttered environment.

\section{Methodology}

Considering that optimization-based planners have the advantage of generating trajectory-level plans suitable for robot execution while satisfying non-trivial constraints, in this work we propose a learning-enhanced trajectory optimization algorithm to address the current research gap and improve both generation speed and solution quality.

\subsection{Problem Statement}

We denote a robot trajectory $\tau$ as a sequence of robot configurations $\theta$ with a fix time interval $\Delta t$ between consecutive states, i.e., $\tau = \left[\theta_0, \theta_1, \ldots, \theta_{K-1} \right] \in \mathbb{R}^{K \times J}$, where $K$ is the trajectory length and $J$ is the dimensionality of the robot configuration. The robot motion planning problem can then be defined as given an initial configuration $\theta_0$, find a trajectory $\tau$ such that the robot reaches the goal configuration $\theta_g$ within a specified distance tolerance. This motion planning problem can be formulated as a time-discretized trajectory optimization:
\begin{equation}
\begin{aligned}
    \arg\min_{\tau} \quad & 
    C_{\text{goal}}(\theta_g, \theta_{K-1}) + C_{\text{smooth}}(\tau) + C_{collision}(\tau) \\
    \text{s.t.} \quad & \theta^{-} \leq \theta_k \leq \theta^{+}, \quad \forall k \in [0,K-1] \\
    & \dot{\theta}^{-} \leq \dot{\theta}_k \leq \dot{\theta}^{+}, \quad \forall k \in [0,K-1] \\
    & \ddot{\theta}^{-} \leq \ddot{\theta}_k \leq \ddot{\theta}^{+}, \quad \forall k \in [0,K-1] \\
    & \dddot{\theta}^{-} \leq \dddot{\theta}_k \leq \dddot{\theta}^{+}, \quad \forall t \in [0,K-1]
\end{aligned}
\end{equation}
where $C_{\text{goal}}(\cdot)$ is a task-related cost that encourages the robot to reach the desired goal configuration within a specified tolerance, $C_{\text{smooth}}(\cdot)$ is a cost term that measures the smoothness of the robot trajectory, $C_{\text{collision}}(\cdot)$ penalizes collisions, including both self-collisions and collisions with obstacles in the environment, and the constraints enforce the physical feasibility of the trajectory by imposing hard limits on joint values, velocities, accelerations, and jerks.

Generally, the objective function in the above optimization problem is highly non-convex, and the quality of the solution is sensitive to the initial guess. A common strategy is to use a linear interpolation between $\theta_0$ and $\theta_g$ as the initial seed for the optimization \cite{toussaint2015logic}. However, when such a seed converges to a local minimum and fails to solve the planning problem, sampling-based algorithms are often used to obtain a better initial guess, at the cost of increased planning time. To address this issue, we propose a learning-enhanced trajectory optimization framework that leverages a novel Flow Matching model as a neural initializer. Our method generates multiple near-optimal seeds simultaneously, significantly improving both the convergence speed and the success rate. The detailed methodology is presented in the remainder of this section.

\begin{figure*}
    \begin{center}
    \includegraphics[width=2.0\columnwidth]{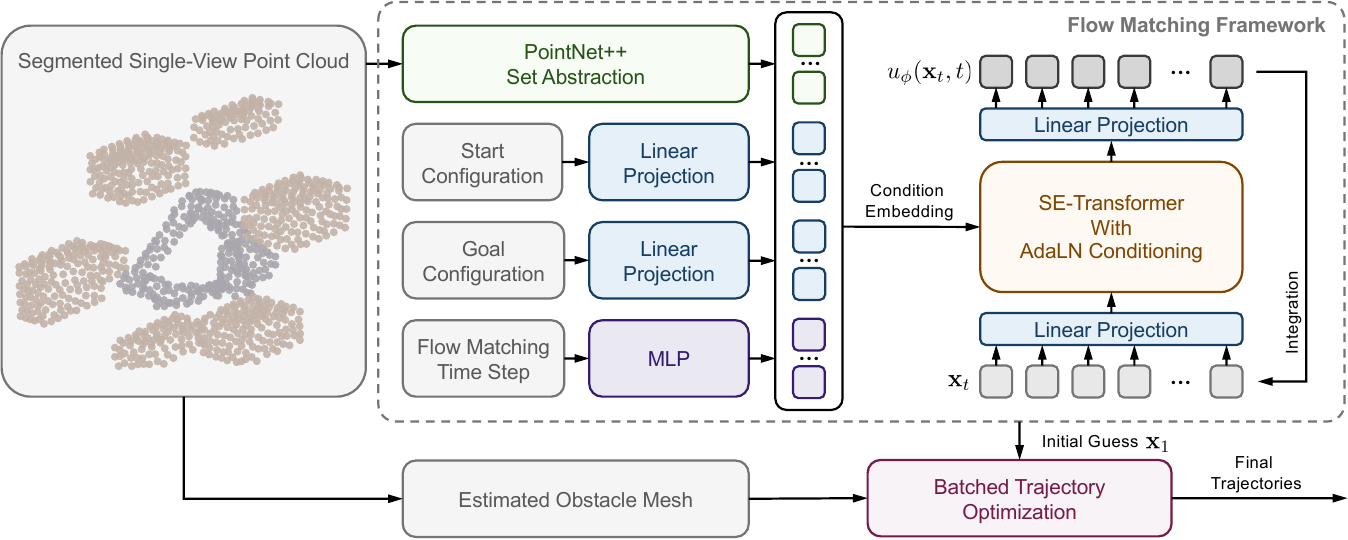}
    \caption{The architecture of proposed learning-enhanced trajectory optimization framework. Our method employs a novel Flow Matching-based neural initializer to generate multiple seeds and passes them to batched optimizer for further refinement. Detailed architecture of SE-Tranformer Block can be seen in \cite{tian2024transfusion}.}
    \label{architecture}
    \end{center}
    \vspace{-0.2in}
\end{figure*}

\subsection{Flow Matching-based Stochastic Neural Initializer}

Flow Matching \cite{lipmanflow} is a recently developed generative model for learning complex probability distributions. It defines a continuous-time flow that transports a simple source distribution $p_0$, often chosen as a standard normal distribution, to a target distribution $p_1$ via a velocity field $u(\mathbf{x}, t)$, which can be parameterized by a neural network $u_{\phi}(\cdot)$. Formally, let $\mathbf{x}_0 \sim p_0$ and $\mathbf{x}_1 \sim p_1$ denote samples from the source and target distributions, and let $\mathbf{x}_t \sim p_t$ represent samples evolving along a transport path at time $t \in [0,1]$. The instantaneous evolution of samples along this path is governed by the velocity field:
\begin{equation}
    \frac{d}{dt} \mathbf{x}_t = u(\mathbf{x}_t, t).
\end{equation}

Flow Matching trains the parameters of $u_{\phi}(\cdot)$ by minimizing the expected squared difference between the predicted velocity and the ground-truth derivative along the transport path, i.e.,
\begin{equation}
    \mathcal{L}_{\text{FM}} (\phi) =\mathbb{E}_{\mathbf{x}_0, \mathbf{x}_1, t} \left[ \| u_{\phi}(\mathbf{x}_t, t) - u(\mathbf{x}_t, t) \|^2 \right].
\end{equation}
In practice, a straight-line transport path is often adopted for more efficient training, faster sample generation, and improved performance. In this case, the intermediate sample $\mathbf{x}_t$ can be expressed as a linear interpolation of $\mathbf{x}_0$ and $\mathbf{x}_1$:
\begin{equation}
    \mathbf{x}_t = (1-t)\mathbf{x}_0 + t \mathbf{x}_1.
\end{equation}
Thus, the loss function for training the velocity field neural network becomes:
\begin{equation}
    \mathcal{L}_{\text{FM}} (\phi) = \mathbb{E}_{\mathbf{x}_0, \mathbf{x}_1, t} \left[ \| u_{\phi}(\mathbf{x}_t, t) - (\mathbf{x}_1 - \mathbf{x}_0) \|^2 \right],
\end{equation}
where $t \sim \mathcal{U}[0,1]$, $\mathbf{x}_0 \sim \mathcal{N}(0,I)$, and $\mathbf{x}_1 \sim p_1$. After training, a sample from the target distribution can be generated by drawing $x_0 \sim \mathcal{N}(0,I)$ and integrating the learned velocity field $u_{\phi}(\cdot)$ along the flow from $t=0$ to $t=1$.

In this work, we propose a novel neural network architecture for $u_{\phi}(\cdot)$, building on the SE-Transformer introduced in \cite{tian2024transfusion}. The robot trajectory representation $\tau \in \mathbb{R}^{K \times J}$ is first projected through an encoding layer for channel lifting, where the embedding of each time step is treated as a token for the Transformer input. Combined with a learnable positional embedding, these tokens are then processed by a stack of SE-Transformer blocks, augmented with a concatenation-based long skip connection to stabilize training. Task-related information, including the start configuration $\theta_0 \in \mathbb{R}^J$, goal configuration $\theta_g \in \mathbb{R}^J$, and the segmented point cloud $P \in \mathbb{R}^{N \times 4}$, is used as a condition to guide sample generation. We use a single-view point cloud that includes both coordinate features and a label feature to provide perceptual information in a partially observable setting. $N$ denotes the total number of points, and the label indicates whether a point belongs to an obstacle or the robot. We employ the PointNet++ Set Abstraction (SA) module \cite{qi2017pointnet++} to extract features from the point clouds. Moreover, we use two linear layers to encode the start and goal configurations separately, and leverage an MLP to encode the Flow Matching time step $t$. After encoding all the conditions, we concatenate the resulting embeddings and utilize adaptive layer normalization (AdaLN) \cite{peebles2023scalable} to modulate the latent variables in the SE-Transformer across multiple layers by regressing the dimension-wise scale $\gamma$, shift $\beta$, and gate $\alpha$ from the condition embeddings. The neural network finally outputs the estimated velocity field through a linear layer following the SE-Transformer blocks, representing the instantaneous velocity of $\mathbf{x}_t$ along the transport path.

\subsection{Learning-enhanced Trajectory Optimization Framework}

The Flow Matching-based neural initializer proposed in the previous section can learn complex robot trajectory distributions in cluttered environments and efficiently sample multiple diverse near-optimal seeds in batch for further refinement. We then employ the state-of-the-art GPU-accelerated optimizer cuRobo \cite{sundaralingam2023curobo} for trajectory optimization. CuRobo is capable of solving multiple optimization problems in parallel, which aligns well with our stochastic neural initializer. CuRobo’s solver first runs a few iterations of particle-based optimization to guide the seed toward a promising region, then applies L-BFGS for rapid convergence. Prior to optimization, we estimate the obstacle mesh from a single-view point cloud, which cuRobo then uses for collision checking during optimization process. Consequently, our framework requires no prior knowledge of obstacle locations or geometries and can generate feasible trajectories directly from depth camera observations. By combining the stochastic neural initializer’s ability to provide diverse seeds with cuRobo’s parallel optimization, the framework is particularly well-suited for dynamic environments where frequent replanning from partial observations is necessary. The detailed architecture of our method can be seen in Fig. \ref{architecture}.

\section{Experiments}

\begin{table*}[htbp]
\caption{Success Rate Comparison with Batched Optimization}
\begin{center}
\resizebox{2.0\columnwidth}{!}{
\begin{threeparttable}
\begin{tabular}{cc|cccc|ccc|cccc|ccc}
\toprule
& & \multicolumn{4}{c|}{Transformer} & \multicolumn{3}{c|}{Transformer-batch} & \multicolumn{4}{c|}{Linear Interpolation} & \multicolumn{3}{c}{Linear Interpolation-batch}  \\ \midrule
Env ID & \# Test Cases & Init. & 5 Iter. & 25 Iter. & 100 Iter. & 5 Iter. & 25 Iter. & 100 Iter. & Init. & 5 Iter. & 25 Iter. & 100 Iter. & 5 Iter. & 25 Iter. & 100 Iter.\\ \midrule
Env 1 & 113 & \textbf{50.4\%} & 65.5\% & 68.1\% & 75.2\% & \textbf{80.5\%} & \textbf{84.1\%} & \textbf{90.3\%} & 41.6\% & 54.9\% & 56.6\% & 59.3\% & 61.9\% & 62.8\% & 64.6\%\\
Env 2 & 134 & \textbf{44.0\%} & 65.7\% & 67.9\% & 75.4\% & \textbf{77.6\%} & \textbf{79.9\%} & \textbf{89.6\%} & 38.8\% & 54.5\% & 55.2\% & 60.4\% & 63.4\% & 63.4\% & 66.4\%\\
Env 3 & 124 & \textbf{41.1\%} & 51.6\% & 59.7\% & 73.4\% & \textbf{66.1\%} & \textbf{78.2\%} & \textbf{88.7\%} & 30.6\% & 53.2\% & 55.6\% & 59.7\% & 61.3\% & 62.9\% & 66.9\%\\
Env 4 & 136 & \textbf{42.6\%} & 55.1\% & 64.7\% & 74.3\% & \textbf{77.9\%} & \textbf{84.6\%} & \textbf{88.2\%} & 25.7\% & 45.6\% & 46.3\% & 49.3\% & 54.4\% & 55.9\% & 59.6\%\\
Env 5 & 126 & \textbf{34.9\%} & 61.1\% & 67.5\% & 75.4\% & \textbf{77.8\%} & \textbf{82.5\%} & \textbf{91.3\%} & 28.6\% & 54.0\% & 54.0\% & 56.3\% & 64.3\% & 65.1\% & 68.3\%\\
Env 6 & 125 & \textbf{49.6\%} & 68.8\% & 73.6\% & 80.0\% & \textbf{83.2\%} & \textbf{87.2\%} & \textbf{94.4\%} & 39.2\% & 66.4\% & 70.4\% & 75.2\% & 76.0\% & 75.2\% & 80.0\%\\ \midrule
Seen (Env 1-6) & 758 & \textbf{43.7\%} & 61.2\% & 66.9\% & 75.6\% & \textbf{77.2\%} & \textbf{82.7\%} & \textbf{90.4\%} & 33.9\% & 54.6\% & 56.2\% & 59.9\% & 63.5\% & 64.1\% & 67.5\% \\ 
Unseen (Env 7) & 634 & \textbf{42.9\%} & 60.7\% & 66.4\% & 75.6\% & \textbf{81.1\%} & \textbf{84.9\%} & \textbf{91.3\%} & 31.9\% & 50.5\% & 51.9\% & 56.0\% & 60.4\% & 61.4\% & 65.8\%\\
\bottomrule
\end{tabular}
\begin{tablenotes}
     \item[*] `Init.' indicates using the initializer alone, while `5/25/100 Iters.' refers to applying the corresponding number of optimization iterations after initialization. The best success rates for initialization and for results after varying numbers of optimization iterations across different environments are highlighted in bold. For the Transformer-batch and Linear Interpolation-batch cases, we duplicate the deterministic result 10 times to run 10 optimizations in batch. The other strategies use a single deterministic seed and produce a single optimized trajectory. The reported success rates are computed using ground-truth obstacles for collision checking to represent true feasibility.
\end{tablenotes}
\label{success_batch}
\end{threeparttable}
}
\end{center}
\end{table*}

\begin{table*}[htbp]
\caption{Success Rate Comparison with Learning-based Initializer}
\begin{center}
\resizebox{2.0\columnwidth}{!}{
\begin{threeparttable}
\begin{tabular}{cc|cccc|cccc|cccc}
\toprule
& & \multicolumn{4}{c|}{FM with One-step Inference (Ours)} & \multicolumn{4}{c|}{FM with Two-step Inference (Ours)} & \multicolumn{4}{c}{Transformer-batch}  \\ \midrule
Env ID & \# Test Cases & Init. & 5 Iter. & 25 Iter. & 100 Iter.  & Init. & 5 Iter. & 25 Iter. & 100 Iter.& Init. & 5 Iter. & 25 Iter. & 100 Iter.\\ \midrule
Env 1 & 113 & 49.6\% & 82.3\% & 89.4\% & 93.8\% & \textbf{55.8\%} & \textbf{89.4\%} & \textbf{90.3\%} & \textbf{94.7\%} & 50.4\% & 80.5\% & 84.1\% & 90.3\%\\
Env 2 & 134 & 45.5\% & 78.4\% & 85.1\% & 90.3\% & \textbf{49.3\%} & \textbf{84.3\%} & \textbf{88.8\%} & \textbf{95.5\%} & 44.0\% & 77.6\% & 79.9\% & 89.6\% \\
Env 3 & 124 & 39.5\% & 71.8\% & \textbf{82.3\%} & \textbf{91.1\%} & \textbf{42.7\%} & \textbf{77.4\%} & 80.6\% & 89.5\% & 41.1\% & 66.1\% & 78.2\% & 88.7\% \\
Env 4 & 136 & 33.8\% & 75.7\% & 79.4\% & 88.2\% & 39.7\% & \textbf{77.9\%} & \textbf{84.6\%} & \textbf{92.6\%} & \textbf{42.6\%} & \textbf{77.9\%} & \textbf{84.6\%} & 88.2\% \\
Env 5 & 126 & 37.3\% & 81.7\% & 86.5\% & 92.9\% & \textbf{42.9\%} & \textbf{86.5\%} & \textbf{90.5\%} & \textbf{95.2\%} & 34.9\% & 77.8\% & 82.5\% & 91.3\% \\
Env 6 & 125 & 43.2\% & 82.4\% & \textbf{88.0\%} & \textbf{94.4\%} & \textbf{49.6\%} & 82.4\% & 85.6\% & 93.6\% & \textbf{49.6\%} & \textbf{83.2\%} & 87.2\% & \textbf{94.4\%}\\ \midrule
Seen (Env 1-6) & 758 & 41.3\% & 78.6\% & 85.0\% & 91.7\% & \textbf{46.4\%} & \textbf{82.8\%} & \textbf{86.7\%} & \textbf{93.5\%} & 43.7\% & 77.2\% & 82.7\% & 90.4\% \\ 
Unseen (Env 7) & 634 & 41.3\% & 82.0\% & 87.1\% & \textbf{93.4\%} & \textbf{45.7\%} & \textbf{86.0\%} & \textbf{89.1\%} & 93.2\% & 42.9\% & 81.1\% & 84.9\% & 91.3\% \\
\bottomrule
\end{tabular}
\begin{tablenotes}
     \item[*] `FM' refers to the Flow Matching-based initializer proposed in this work. `Init.' indicates using the initializer alone, while `5/25/100 Iters.' refers to applying the corresponding number of optimization iterations after initialization. The best success rates for initialization and for results after varying numbers of optimization iterations across different environments are highlighted in bold. For each planning problem, our method generates 10 initial guesses in parallel, followed by 10 optimizations in batch. For the Transformer-batch case, we duplicate the deterministic result 10 times to run 10 optimizations in batch for a fair comparison. The reported success rates are computed using ground-truth obstacles for collision checking to represent true feasibility.
\end{tablenotes}
\label{success_our}
\end{threeparttable}
}
\end{center}
\end{table*}

\subsection{Data Collection}

We generate our dataset using the challenging planning environments presented in \cite{soleymanzadeh2025simpnet}. In the SIMPNet dataset, a motion planning problem is defined by the geometries and positions of multiple obstacles densely arranged within the workspace, along with a specified start-goal configuration pair. Since SIMPNet focuses on path planning problem, we smooth the ground-truth robot path in SIMPNet dataset using B-splines and feed the resulting trajectory into cuRobo for optimization, making it feasible for the robot to execute. Planning problems for which cuRobo fails to find a feasible solution within a specified number of iterations are discarded. To produce partial observations, we reconstruct the planning scene in simulation, including both obstacles and the robot’s initial pose, and place six virtual cameras with randomly sampled positions and orientations within a specified range so that each camera can detect the robot. Depth images are then rendered, and the corresponding extrinsic and intrinsic camera parameters in robot base frame are recorded for each planning problem. We generate 4,416 planning problems with feasible solutions $\tau \in \mathbb{R}^{128 \times 6}$ across seven scenarios with different obstacle arrangements. Each trajectory has a length of 128, and the robot has six degrees of freedom. We reserve one scenario solely for validation in unseen environments. For the remaining six scenarios with different obstacle arrangements, 80\% of the problems are used for training, and 20\% for validation. In summary, we generate 3,024 samples for training, 758 samples for validation in seen environments, and 634 samples for validation in unseen environments.

\subsection{Baselines and Evaluation Metrics}

\textbf{Baselines:} We compare our model with two initialization strategies: linear interpolation and a Transformer-based deterministic neural initializer with an encoder-decoder architecture inspired by \cite{yang2025deep}. Deterministic methods produce only a single initial guess, which after optimization yields one refined trajectory. We refer to these two baselines as Transformer and Linear Interpolation. 

\textbf{Extended Baselines:} Additionally, since cuRobo can solve multiple optimizations in batch and particle-based optimization introduces stochasticity even when the initializations are identical, we extend baselines and report results from batched optimization using the same Transformer-based and linear interpolation-based initializer, thereby highlighting the benefits of stochasticity in trajectory optimization. Specifically, the outputs of these deterministic initializers are replicated multiple times and fed into the cuRobo batched solver to get different optimized trajectories. We refer to these two extended baselines as Transformer-batch and Linear Interpolation-batch.

\textbf{Evaluation Metrics:} We evaluate our methods using two metrics: (1) Planning Time, which measures the time required for the initializer and optimization solver to find a solution; and (2) Success Rate, defined as the percentage of motion planning problems that are successfully solved. A solution is considered feasible if the resulting trajectory reaches the goal, is collision-free, and satisfies the joint limits. We use ground truth obstacles for collision checking to show the true feasibility. Notably, for stochastic trajectory optimization, including our proposed framework and batched optimization with identical seeds, a problem is considered successful if at least one candidate trajectory is feasible \cite{carvalho2023motion}.

\subsection{Implementation Details}

We train the model for 1,000 epochs with an initial learning rate of $3 \times 10^{-4}$ and a batch size of 64. Training is warmed up for 100 epochs before applying a cosine annealing learning rate decay strategy. The model uses a 9-layer SE-Transformer with a latent dimension of 512. For each planning problem, we sample 1,024 points from the segmented depth image to form a point cloud representing the obstacles and the robot. The dimension of the point cloud embedding after PointNet++ is set to 256, the start and goal embeddings are both set to 64, and the Flow Matching time-step embedding has a dimension of 128. During the inference, Flow Matching initializer generates samples in 2 or 1 inference steps. We set the number of initial seeds to 10 for both our proposed model and the extended baselines for fair comparison. That is, our Flow Matching model generates 10 diverse seeds and runs optimization in batch, while Transformer-batch and Linear Interpolation-batch replicate their deterministic seed 10 times and run the batched optimization. Training is performed on a single NVIDIA A100 GPU, and all experimental tests are conducted on a single NVIDIA RTX 4080. Baselines are fine-tuned for optimal performance.

\begin{figure*}
    \begin{center}
    \includegraphics[width=1.85\columnwidth]{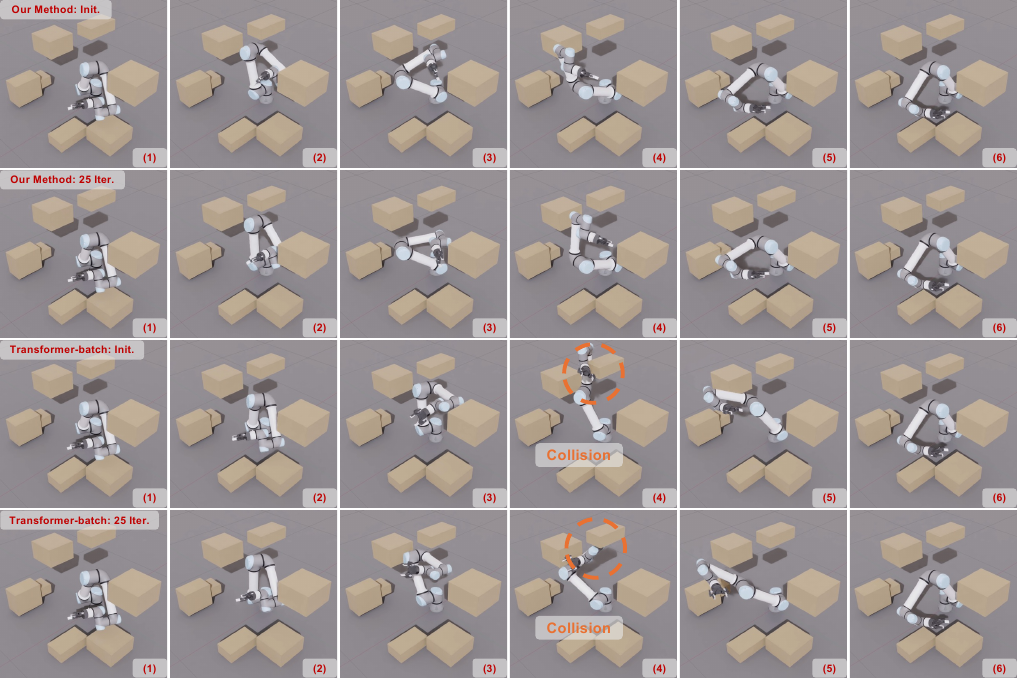}
    \caption{The first simulation case study comparing our method with Transformer-batch. Our method produces feasible initializations and trajectories after 25 optimization iterations, whereas Transformer-batch results in collisions for both the initializations and the optimized trajectories.}
    \label{simulation1}
    \end{center}
    \vspace{-0.2in}
\end{figure*}

\begin{figure*}
    \begin{center}
    \includegraphics[width=1.85\columnwidth]{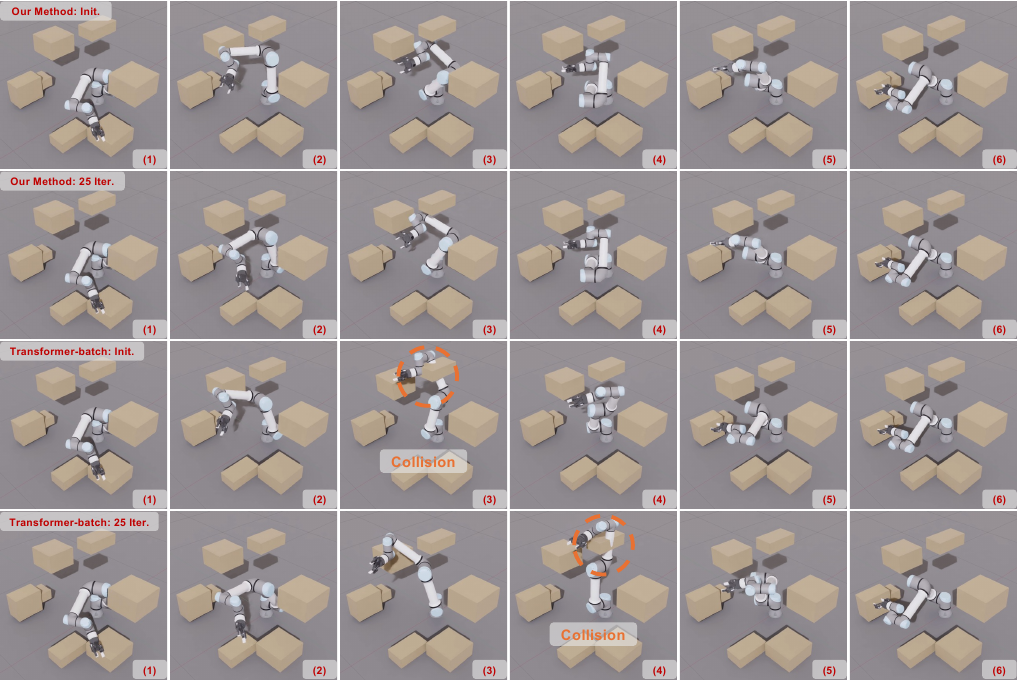}
    \caption{The second simulation case study comparing our method with Transformer-batch. Our method produces feasible initializations and trajectories after 25 optimization iterations, whereas Transformer-batch results in collisions for both the initializations and the optimized trajectories.}
    \label{simulation2}
    \end{center}
    \vspace{-0.2in}
\end{figure*}

\subsection{Comparison Results}

We present the success rate comparison in two tables. Table \ref{success_batch} first compares the Transformer-based deterministic initializer with traditional linear interpolation, as well as with our extended baselines that incorporate the batched optimization solver. As shown in Table \ref{success_batch}, the Transformer achieves a higher success rate than linear interpolation across all cases, indicating that a neural network learning a good initial guess can enhance trajectory optimization performance. Moreover, when comparing our extended baselines, Transformer-batch performs better than the Transformer strategy, and Linear Interpolation-batch performs better than the linear interpolation strategy, indicating that adding stochasticity through batched optimization can enhance the success rate. However, as batched optimization with identical initial seeds only explores uncertainty within a single data mode, it cannot escape local minima if the initial guess is poor. As a result, Transformer-batch achieves a substantially higher success rate than Linear Interpolation-batch due to a good initial guess.

To demonstrate the effectiveness of our proposed stochastic neural initializer, we compare our method to the extended baseline with the best performance, as shown in Table \ref{success_our}. Since Flow Matching learns a continuous near-straight transportation, the integration step can be chosen during inference. In general, sample quality improves with more generation steps at the cost of increased inference time. In this work, we report results for two variants: a one-step inference model and a two-step inference model. For a fair comparison, all models perform a total of ten optimizations in batch for each planning problem. Our model demonstrates a significant overall improvement in planning success rate: specifically, our two-step method with just five optimization iterations can outperform Transformer-batch with 25 iterations and achieves improvements of up to 5\% in both seen and unseen environments. Even our one-step model surpasses Transformer-batch after applying optimization in both seen and unseen scenarios. These improvements indicate that learning a diverse set of near-optimal seeds benefits trajectory optimization, as diverse initialization allows the optimizer to find minima across different data modes rather than exploring a single mode. Although the success rate of our proposed initializer alone is not as high as that achieved after optimization, collisions are usually very shallow and easily corrected. Additionally, the success rate in unseen environments is comparable to that in seen environments, indicating that our method has strong generalization ability.

We further compare the planning time and show the results in Table \ref{time}. The reported times are averaged over more than one hundred problems. The planning time of the Transformer-batch strategy falls between those of our one-step and two-step models. Therefore, our one-step model should generally be preferred over Transformer-batch strategy, as it achieves better performance in both seen and unseen cases after optimization while also offering faster planning times. For applications requiring a higher success rate without strict inference time constraints, our two-step model can be employed. We also present visualizations of the planning results in Fig. \ref{simulation1} and Fig. \ref{simulation2}. In both cases, our method generates feasible solutions both from the initializer alone and after 25 optimization iterations. In contrast, Transformer produces poor initializations in both cases, and even with batched optimization, the algorithm remains trapped in local minima.

\begin{table}[htbp]
\caption{Planning Time Comparison}
\begin{center}
\vspace{-0.15in}
\resizebox{1.0\columnwidth}{!}{%
\begin{threeparttable}
\begin{tabular}{c|cccc}
\toprule
& \multicolumn{4}{c}{Planning Time (Second)} \\ \cmidrule{2-5}
Method & Init. & 5 Iter. & 25 Iter. & 100 Iter.\\ \midrule
FM with One-step Inference (Ours) & 0.06792 & 0.08030 & 0.1136 & 0.2334 \\
FM with Two-step Inference (Ours) & 0.1308 & 0.1432 & 0.1765 & 0.2963 \\
Transformer & 0.07372 & 0.08552 & 0.1181 & 0.2349 \\
Transformer-batch & 0.07372 & 0.08610 & 0.1194 & 0.2392 \\
Linear Interpolation & 0.0001659 & 0.01135 & 0.04451 & 0.1614 \\
Linear Interpolation-batch & 0.0001659 & 0.01255 & 0.04589 & 0.1657 \\
\bottomrule
\end{tabular}
\begin{tablenotes}
     \item[*] `FM' refers to the Flow Matching-based initializer proposed in this work. `Init.' indicates using the initializer alone, while `5/25/100 Iters.' refers to applying the corresponding number of optimization iterations after initialization.  The unit of planning time is seconds. The reported results are averaged over one hundred planning problems.
\end{tablenotes}
\end{threeparttable}
}
\vspace{-0.2 in}
\end{center}
\label{time}
\end{table}

\section{Conclusion}

Trajectory optimization is sensitive to the initial guess and can easily converge to local minima in obstacle-dense environments. In this work, we introduce a novel Flow Matching-based initializer that simultaneously generates multiple diverse near-optimal seeds to warm-start trajectory optimization. Combined with a state-of-the-art batched trajectory optimization solver, our method can explore minima across different data modes. Our framework generates robot trajectories without requiring knowledge of obstacle locations or geometries, making it well-suited for fast replanning in dynamic environments. Compared with the commonly used linear interpolation strategy and a learning-based approach, our method improves the success rate of trajectory optimization and also accelerates convergence.

\bibliographystyle{IEEEtran}
\bibliography{ref}{}

\end{document}